\documentclass{bmvc2k}

\usepackage{pdfpages}
\usepackage{multirow}
\usepackage{booktabs}
\usepackage{wrapfig}
\usepackage{amssymb}
\usepackage{algpseudocode}
\usepackage{algorithm}
\usepackage{setspace}

\title{Free-form 3D Scene Inpainting\\ with Dual-stream GAN}

\addauthor{Ru-Fen Jheng}{ruby2332ruby@cmlab.csie.ntu.edu.tw}{1}
\addauthor{Tsung-Han Wu}{tsunghan@cmlab.csie.ntu.edu.tw}{1}
\addauthor{Jia-Fong Yeh}{jiafongyeh@ieee.org}{1}
\addauthor{Winston H. Hsu}{whsu@ntu.edu.tw}{1,2}
\addinstitution{
 National Taiwan University
}
\addinstitution{Mobile Drive Technology}

\runninghead{Jheng et al.}{Free-form 3D Scene Inpainting with Dual-stream GAN}

\def\ourdataset{FF-Matterport}
\newcommand{\ts}{\textsuperscript}

\begin{document}

\maketitle

\begin{abstract}
Nowadays, the need for user editing in a 3D scene has rapidly increased due to the development of AR and VR technology. However, the existing 3D scene completion task (and datasets) cannot suit the need because the missing regions in scenes are generated by the sensor limitation or object occlusion. Thus, we present a novel task named free-form 3D scene inpainting. Unlike scenes in previous 3D completion datasets preserving most of the main structures and hints of detailed shapes around missing regions, the proposed inpainting dataset, FF-Matterport, contains large and diverse missing regions formed by our free-form 3D mask generation algorithm that can mimic human drawing trajectories in 3D space. Moreover, prior 3D completion methods cannot perform well on this challenging yet practical task, simply interpolating nearby geometry and color context. Thus, a tailored dual-stream GAN method is proposed. First, our dual-stream generator, fusing both geometry and color information, produces distinct semantic boundaries and solves the interpolation issue. To further enhance the details, our lightweight dual-stream discriminator regularizes the geometry and color edges of the predicted scenes to be realistic and sharp. We conducted experiments with the proposed FF-Matterport dataset. Qualitative and quantitative results validate the superiority of our approach over existing scene completion methods and the efficacy of all proposed components. Our code is available at \href{https://github.com/ruby2332ruby/Free-form-3D-Scene-Inpainting}{https://github.com/ruby2332ruby/Free-form-3D-Scene-Inpainting}
\end{abstract}

%-------------------------------------------------------------------------

% \input{final_sec/1_Introduction}
% \input{final_sec/2_RelatedWork}
% \input{final_sec/3_Method}
% \input{final_sec/4_Experiments}
% \input{final_sec/5_Conclusion}
\section{Introduction}
\label{sec:intro}

\begin{figure*}[t]
\begin{center}
\includegraphics[scale=0.305]{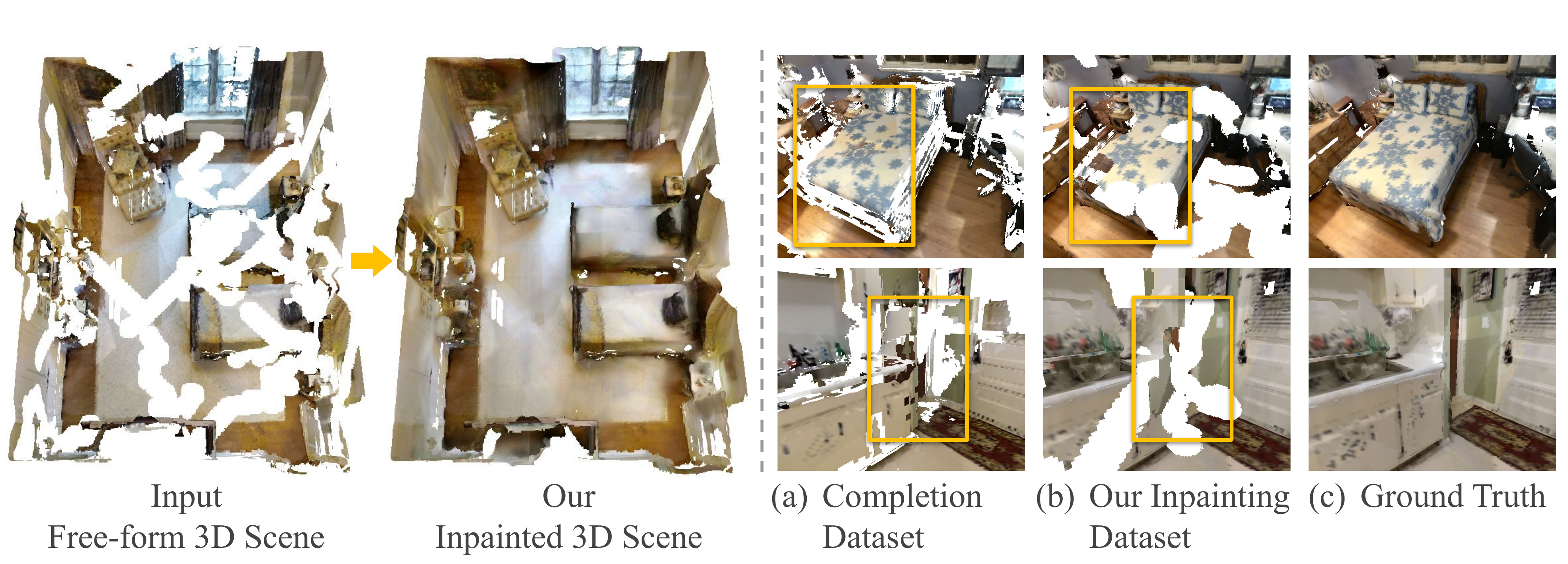}
\vspace{-1.1cm}
\end{center}
   \caption{{\bf Left:} We propose a novel 3D scene inpainting task with the first free-form 3D scene dataset,~\ourdataset{}, which imitates user drawing masks in 3D space. Our proposed model takes the incomplete scene as input and recovers the missing parts with high-quality and realistic results. {\bf Right:} Compared with the prior scene completion dataset generated by incomplete observations and preserving main structures and hints of missing parts, our dataset contains challenging yet practical missing regions for the 3D scene inpainting task.}
\label{fig:fig1}
\end{figure*}

In recent years, Augmented Reality (AR) and Virtual Reality (VR) have become popular in our daily life, such as VR gaming, virtual tours, and AR meeting software. To form realistic 3D scenes in these applications, reconstructing scenes from multiple sensed RGB-D images is a widely-used and cost-effective approach. However, users often want to further edit the reconstructed 3D scenes to meet their needs. Take, as an example, removing unwanted objects existing in the real world. Therefore, there is still a strong need for 3D post-processing.

In this work, we introduce this application as the 3D scene inpainting task. Specifically, given a 3D scene with several manually specified 3D masks, a 3D inpainting model should fill these regions with proper contents, including both geometry and color. Although 3D completion methods~\cite{song2017semantic,dai2018scancomplete,dai2020sg,dai2021spsg} also aim to complete missing regions in 3D space, they cannot meet the needs of the 3D inpainting application for the following reasons:

{\bf Insufficient Evaluation:} The existing 3D scene completion datasets~\cite{dai2020sg,dai2021spsg} are unsuitable for the inpainting task because they (1) lack the masks that specify the missing regions and (2) form the missing regions only by sensor limitation or object occlusion. About (1), regions with a 0-value could be either missing parts that need to be repaired or an empty space without objects due to the sparseness property of 3D space. Without the informative masks of missing regions, SOTA methods in 3D completion thus change the shape of completed parts or leave the incomplete regions with small holes or artifacts (Fig.~\ref{fig:result}). Regarding (2), the physical properties of missing regions in 3D inpainting tasks and existing 3D scene completion datasets~\cite{dai2020sg,dai2021spsg} differ. To be specific, incomplete areas in the existing completion datasets are due to inherent sensor limitation or object occlusion. They are usually regular and strongly correlated to object occlusion or specific camera view (Fig.~\ref{fig:fig1}). In contrast, the missing regions in 3D inpainting tasks are irregularly shaped and could randomly occur anywhere in a scene. To effectively validate the methods in the 3D inpainting task, we propose a novel {\it Free-Form Matterport3D (FF-Matterport) dataset} tailored for this task. Free-form masks imitate the diverse human drawing trajectories in real-case (Sec.~\ref{sec:M_data}).

{\bf Poor Geometry and Color Reconstruction:} As aforementioned, previous methods lack the ability to recover the missing regions well in 3D inpainting tasks. First, they cannot attend to the missing parts due to the lacking of mask information. To tackle this, we leverage the mask information via modifying the gated convolution module~\cite{yu2019free} to guide our model to focus on the crucial area. Second, the SOTA method~\cite{dai2021spsg} recovers missing areas with a single-stream two-stage generator, which generates geometry structure (first-stage) and color (second-stage) sequentially. Since the unstable geometry results learned in the first stage are without the help of semantic features from color, it may cause the error propagation issue resulting in poor inpainting results. Observing this, we introduce the first {\it 3D dual-stream generator} (Fig.~\ref{fig:model} (a)) to collaboratively generate both geometry and color in missing regions (Sec.~\ref{sec:M_dual_g}). By considering the information of two modalities simultaneously, our dual-stream generator produces more realistic object structures than the single-stream generator.

{\bf Crude Details:} As the 3D scene inpainting task is much more challenging, prior approaches show the further shortcoming, i.e., producing crude details of restoration. In brief, prior colored 3D completion work~\cite{dai2021spsg} leveraged a color discriminator to make the rendered images of the generated scene similar to the real scene with adversarial loss. However, this color adversarial discriminator still results in blurred and distorted boundaries when recovering large damaging regions with structural contents, such as picture frames or furniture edges. To enhance the sharpness and structure of boundaries, we introduce an extra edge stream discriminator apart from the color adversarial discriminator. Particularly, we randomly project the inpainted 3D scene to multiple 2D images and then simultaneously constrain its color texture along with the predicted edge maps using our {\it dual-stream discriminator} (Fig.~\ref{fig:model} (b) and Sec.~\ref{sec:M_dual_d}). With additional regularizations on the edges of global structures, we find that our geometry and color generators learn to collaborate and generate less blurry and well-structured boundaries.

We validate our dual-stream GAN and previous 3D scene completion methods on the proposed FF-Matterport dataset. Our method demonstrates the superiority in six different metrics. Besides, the impressive visualization results also illustrate the effectiveness of our method. To sum up, this work presents the following main novelties and contributions:

\begin{itemize}
    \item We propose a novel 3D scene inpainting task with the first free-form 3D scene dataset, FF-Matterport, which contains diverse free-form masks generated with our designed algorithm imitating human drawing trajectories.
    \item We introduce the first 3D gated dual-stream (geometry and color) generator to jointly consider the geometry and color context of missing regions and generate high-quality contents with semantic-constrained structures.
    \item We introduce an edge and color dual-stream discriminator guiding the generator to produce clear and detailed geometry and color boundaries.
\end{itemize}
\section{Related Work}
\label{sec:rela}

\subsection{3D Completion}
\label{sec:SC}

In the field of 3D vision, object completion is a fundamental and long-standing problem. Some took single or multi images to reconstruct or complete 3D objects~\cite{choy20163d,mescheder2019occupancy,saito2019pifu, sun2018im2avatar,xu2019disn,groueix2018papier}. Others utilized depth or RGB-D frames collected by commodity depth sensors to reconstruct 3D objects~\cite{wu20153d,wu2018learning}. Still, others aimed to complete a 3D object by various 3D representations, such as 3D point cloud scan~\cite{han2017high,yang2017foldingnet,yuan2018pcn,sarmad2019rl,huang2020pf,yu2021pointr}, sign distance fields (SDF)~\cite{dai2017shape,park2019deepsdf}, or mesh surface~\cite{kazhdan2006poisson,kazhdan2013screened}. Nonetheless, the above methods are limited to 3D objects rather than a complex 3D scene with several items. 

To complete a complicated 3D scene, some recent studies have been proposed.~\cite{newcombe2011kinectfusion,izadi2011kinectfusion,dai2017bundlefusion,huang20173dlite,huang2020adversarial,maier2017intrinsic3d, roldao20223d,li2022compnvs,zhang2019cascaded,wang2020deep,li2020anisotropic} leveraged scanned multiple RGB-D images to reconstruct and complete a 3D scene. SSCNet~\cite{song2017semantic} combined the scene completion task with the 3D semantic segmentation task. ScanComplete~\cite{dai2018scancomplete} extended the 3D completion task to large scenes and designed the network to handle various scene scales during inferring. SG-NN~\cite{dai2020sg} first trained and evaluated the 3D scene completion model on a real-world scanned dataset, and SPSG~\cite{dai2021spsg} first tackled color completion in 3D scenes apart from geometry completion. However, the above prior works only focus on completing missing parts due to sensor limitations, which are strongly correlated to viewing angles and have generally similar patterns.

To the best of our knowledge, we are the first to introduce the 3D scene inpainting task and generate a free-form 3D scene dataset to train and evaluate the performance of completing manual masks in 3D space. Furthermore, our dual-stream GAN, complementing geometry and color information with each other, solves the problem of over-smoothed geometry shapes and blurred color boundaries in prior single-stream two-stage work~\cite{dai2021spsg}.

\subsection{2D Image Inpainting}

2D image inpainting takes a corrupted image as input and fills the missing parts in the image with semantically correct, and boundary-consistent contents. It is an important task for many downstream visual tasks, such as object removal, damaged photo restoration, and 2D to 3D photo transformation~\cite{shih20203d}. Traditional approaches~\cite{barnes2009patchmatch,efros2001image,darabi2012image,huang2014image} reused the patches from the image background or source images to repair the missing pixels with the most similar one, but they only can handle repetitive patterns or small missing holes. 

Recently, GAN~\cite{goodfellow2014generative} has made great progress in the image inpainting task, enabling inpainting models to fill holes with realistic and semantically reasonable contents \cite{denton2015deep}. PConv~\cite{liu2018image} and GatedConv~\cite{yu2019free} extended regular rectangle masks to free-form masks with irregular shapes and developed corresponding CNN modules to handle the more challenging masks. More recently, some practices~\cite{nazeri2019edgeconnect,liu2020rethinking,guo2021image} utilized additional edge constraints to conquer the blurry results on large missing areas where the main structure of the object is missing.~\cite{nazeri2019edgeconnect} proposed an edge-color two-stage inpainting framework;~\cite{liu2020rethinking} and~\cite{guo2021image} developed new generator models to combine and exchange structure and texture information. Some others extended image inpainting to multi-view~\cite{mori2020inpaintfusion,thonat2016multi,jampani2021slide,philip2018plane} or light field~\cite{le2018light}.

Different from prior 2D inpainting studies, we first introduce the 3D inpainting task and present a 3D-specific free-form mask generation algorithm due to the sparseness property in 3D space. Moreover, inspired by \cite{nazeri2019edgeconnect,liu2020rethinking,guo2021image}, we propose the first dual-stream GAN for the 3D inpainting task, not only cooperating 3D geometry and color information in the generator but also regularizing color and edges from diverse viewpoints in the discriminator.

\begin{figure*}[!t]
\begin{center}
\includegraphics[scale=0.228]{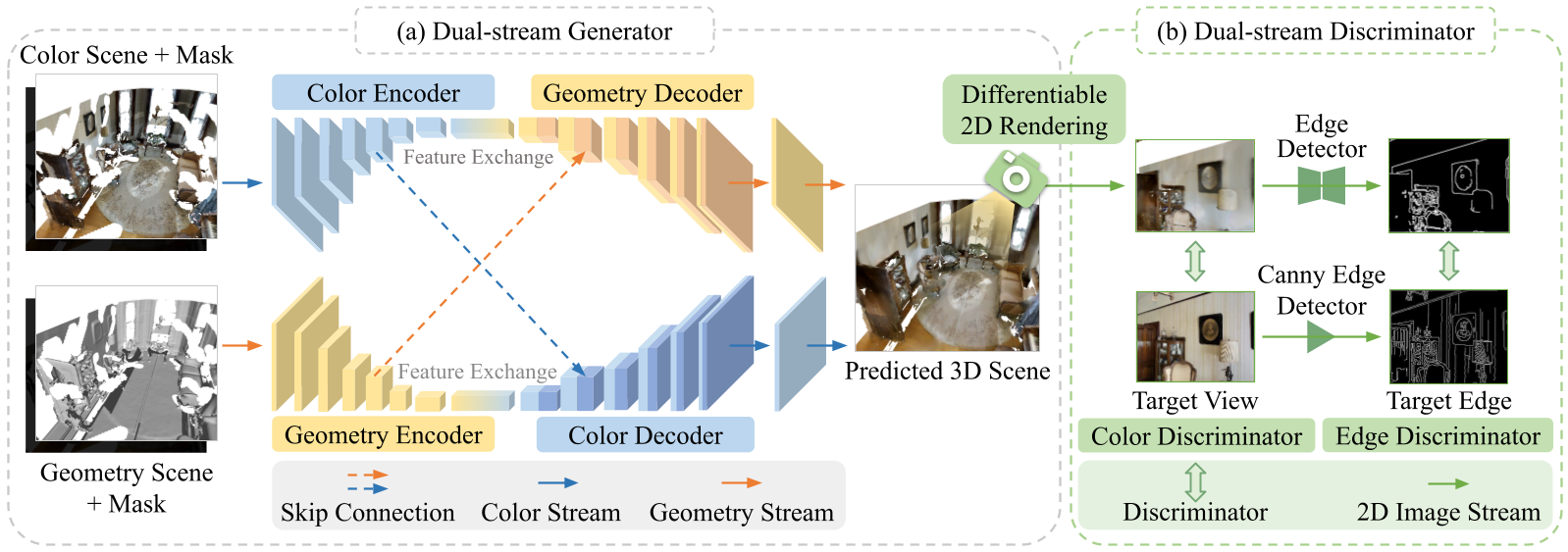}
\vspace{-1.2 cm}
\end{center}
\caption{Overview of the proposed network. (a) Geometry and color dual-stream generator exchanges and fuses embedded features from each other to complement respective decoders. (b) The dual-stream discriminator uses the differentiable 2D rendering and the edge detector to project the predicted scene to a 2D image and an edge map. Then, it optimizes them with the target view and corresponding canny edge image.}
\vspace{-0.6 cm}
\label{fig:model}
\end{figure*}

% \vspace{-0.5cm}
\section{Method}
\label{sec:method}
\vspace{-0.2cm}

In this work, we propose a novel 3D scene inpainting task with the tailored FF-Matterport dataset for the need in AR and VR applications. The 3D scenes in the dataset contain irregular and diverse shapes of missing regions (masks) generated by our novel free-form 3D mask generation algorithm (Sec.~\ref{sec:M_data}) that mimics humans drawing trajectories in 3D scenes. To tackle this challenging yet practical task, we develop a dual-stream GAN model (Fig.~\ref{fig:model}) that contains two main components, the dual-stream generator and the dual-stream discriminator. The dual-stream generator (Sec.~\ref{sec:M_dual_g}) leverages the mask information and the feature fusion of geometry and color to generate semantic-constraint structures and shape-constraint textures. In addition, the dual-stream discriminator (Sec.~\ref{sec:M_dual_d}) further enhances the details by regularizing color and corresponding edges on randomly rendered images. Finally, we summarize the overall training loss. (Sec.~\ref{sec:M_train_loss})

\vspace{-0.2cm}
\subsection{Free-form 3D Dataset Generation} %include problem and input definition
\label{sec:M_data}

\begin{wrapfigure}{r}{0.55 \linewidth}
\vspace{-0.2cm}
\includegraphics[scale=0.228]{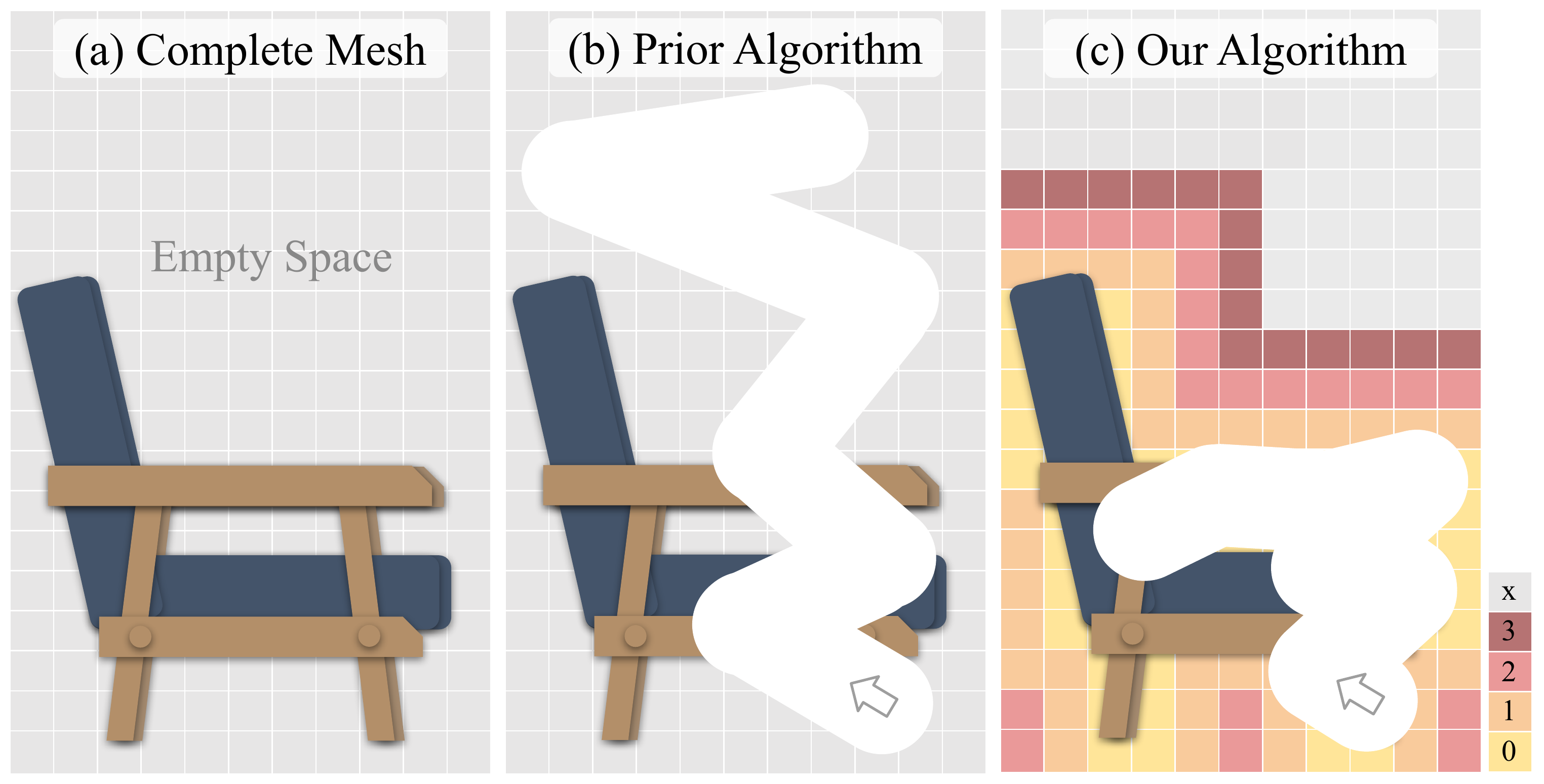}
\caption{Trajectories comparison between (b) previous algorithm and (c) our algorithm when drawing in (a). Due to the sparseness of 3D space, algorithm (b)~\cite{yu2019free} generates masks on empty space and remains weird object shapes. In contrast, our free-form 3D algorithm (c) is able to produce practical masks around the surface with better flexibility by utilizing the property of TSDF representation.}
\vspace{-0.2cm}
\label{fig:sdf}
\end{wrapfigure}

As illustrated in Fig.~\ref{fig:fig1}, our free-form 3D dataset aims to alleviate the issues of too regular patterns in missing areas and the lacking mask information in existing 3D completion datasets~\cite{dai2020sg,dai2021spsg}. To generate free-form 3D masks, it is intuitive to modify the free-form 2D mask generation algorithm~\cite{yu2019free}, which uses strokes with random lengths and directions to line up a drawing track. As all pixels in 2D images contain information, drawing straight lines back and forth can cover the unwanted objects. However, since objects in 3D space are sparse and have curved and complicated shapes, directly applying the previous algorithm in 3D space usually masks areas without objects or produces meaningless shapes (Fig.~\ref{fig:sdf}).

To conquer the above challenges, we design a novel 3D mask generation algorithm. Initially, it converts the 3D scene data to the truncated signed distance field (TSDF) representation. The algorithm can thus ensure that the painted stroke persists around the surface by checking the TSDF values. Afterward, we use an incremental masking strategy rather than the original 2-point line strategy in~\cite{yu2019free}. Our strategy dynamically decides the direction and length of strokes, considering the distance to the surface and the diameter of the stroke. Besides, we randomly sample points in 3D scenes as the starting point for strokes to ensure diversity. With our algorithm, the curved strokes can fit various object shapes and occur in diverse places in 3D scenes. More details are reported in supplementary materials.

%We apply our mask generation algorithm to the Matterport3D~\cite{chang2017matterport3d} dataset with the official train-test split and produce the first free-form 3D scene inpainting dataset, named \ourdataset{}. It contains 30-40\% missing regions randomly located in the whole indoor scene and is voxelized to 2cm resolution same as~\cite{dai2020sg,dai2021spsg}. Notably, unlike in the 2D image inpainting task where a mask can be randomly paired with any images, each free-form 3D mask in our dataset is generated according to the object distribution of its corresponding scene.

% 10/04 Ruby
We apply our mask generation algorithm to the Matterport3D~\cite{chang2017matterport3d} dataset with the official train-test split and produce the first free-form 3D scene inpainting dataset, named \ourdataset{}. It contains 30-40\% missing regions randomly located in the whole indoor scene and is voxelized to 2cm grids (same resolution as~\cite{dai2020sg,dai2021spsg}), where we stored fused TSDF, RGB value, and binary mask in separate channels. Notably, unlike in the 2D image inpainting task where a mask can be randomly paired with any images, each free-form 3D mask in our dataset is generated according to the object distribution of its corresponding scene.

\subsection{Dual-stream Generator}
\label{sec:M_dual_g}

%As mentioned in Sec.~\ref{sec:M_data}, the scenes in the 3D inpainting task contain irregular and various missing regions, which makes the task more challenging. Moreover, the mask information is also provided to indicate areas that are needed to focus. To better leverage the mask information and reconstruct sophisticated and realistic object surfaces in missing regions, we develop a dual-stream generator specifically for this task.

As previously mentioned, the 3D inpainting task is more challenging because it contains scenes with irregular and various missing regions. To reconstruct sophisticated and realistic object surfaces in these scenes, we need to better leverage all the information from geometry, color, and mask; thus, we develop a dual-stream generator specifically for this task.

% 10/06
%As mentioned in Sec.~\ref{sec:M_data}, the scenes in the 3D inpainting task contain irregular and various missing regions, which makes the task more challenging. To reconstruct sophisticated and realistic object surfaces in these scenes, we need to better leverage all the information from geometry, color, and mask; thus, we develop a dual-stream generator specifically for this task.

To utilize the masks, we can treat them as additional channel inputs. Nonetheless, we further exploit the benefits of masks, attaching a 3D gated convolution module (3D GatedConv) extended from \cite{yu2019free} to the generator. Specifically, the 3D GatedConv module helps the generator gradually fill the masked regions with proper geometry and color contents by dynamically learning soft attention maps. This modification contributes significantly to model performance, and we found that baselines without the mask information or 3D GatedConv module only produce distorted restoration results (See Tab.~\ref{Tab:ablation}).

Regarding the pipeline of the generator, SPSG~\cite{dai2021spsg} developed a two-stage pipeline, which first completes the geometry of all missing areas and then generates color on the surfaces specified in the first stage. We find that the geometry generator in this pipeline completely ignores the semantic features of color. Also, the unstable geometry outcome of the first stage causes error propagation to the second stage. To this end, we argue that the geometry and color information should be considered simultaneously and can benefit each other. Inspired by~\cite{guo2021image}, we develop the first 3D dual-stream (geometry and color) generator that can fuse and retrieve the knowledge from both streams during generation. 

In Fig.~\ref{fig:model} (a), our dual-generator consists of two generators, each with a U-Net variant. In the encoding phase, the geometry and color scenes are embedded independently and projected to high-level feature space through corresponding generators. During the decoding phase, the two generators fuse feature embedding from each other as an additional condition to refine the respective decoded results. Besides, we combine the encoder and decoder features with skip connections to create more delicate content. This operation allows us to fully exchange geometry and color information during generation, producing both semantic-aware geometry structure and shape-constrained color texture in predicted scenes. Compared with the previous pipeline using a one-way feature stream forward from geometry to color, our dual-stream pipeline provides a mutual feature exchange between the two generators. Consequently, we can observe that our pipeline alleviates the error propagation problem, revealing consistent performance improvements in both geometry and color results (Tab.~\ref{Tab:ablation}). 

\subsection{Dual-stream Discriminator}
\label{sec:M_dual_d}

% Motivation
To make the generator produce high-quality objects on missing regions, it is a common practice to directly regularize generated scenes by designing loss functions or utilizing a discriminator. In the field of 3D completion, \cite{dai2020sg} used a naive $\ell1$ loss to regress the geometry outputs and \cite{dai2021spsg} applied a 2D discriminator to force the color outputs realistic under diverse rendering views. Nevertheless, the above practices are inadequate to meet the need of 3D scene inpainting as its missing areas, unlike small corrupted strips due to sensor limitation in 3D completion, are generally large and lack structural contents. Specifically, we found those methods are prone to produce over-smoothed structures and blurred color edges by interpolating nearby color and geometry values or linking mesh pieces (Fig.~\ref{fig:result}). Thus, we develop a novel dual-stream discriminator to avoid producing such crude details.

% Dual-stream discriminator: Method
As illustrated in Fig.~\ref{fig:model} (b), our dual-stream discriminator is composed of two components: a color discriminator and an edge discriminator. To begin with, our color discriminator regularizes the quality of generated scenes on randomly rendered 2D images following~\cite{dai2021spsg}. This 2D color stream makes the projected 2D frames of the generated 3D scene as realistic as possible and is more effective than applying a discriminator in 3D~\cite{dai2021spsg}. Moreover, to ensure fine-grained geometry shapes and sharp color boundaries in the generated scene, we design an edge discriminator further regularizing the corresponding edges of the rendered frames. To elaborate, our edge discriminator compares the 2D edge maps extracted from real projected frames by the Canny edge detector~\cite{canny1986computational} and that extracted from generated frames by our NN-based edge detector. With the aid of our lightweight edge discriminator, we can not only produce sharp color boundaries and detailed geometry contents on qualitative results (Fig.~\ref{fig:result}) but also achieve huge improvements in quantitative results (Tab.~\ref{Tab:ablation}).

% Summary
%To summarize, our training objectives can be categorized into two groups. For the naive full supervision loss, we supervise our geometry and color reconstruction from both 3D and 2D space. The $L_{geo}$ following the log-transformed $\ell1$ TSDF loss in~\cite{dai2020sg} is applied to penalize the geometry in 3D space, and the $L_{color}$ as well as depth $L_{depth}$ (both are $\ell1$ losses) are leveraged to ensure geometry and color reconstruction on the rendered images. Furthermore, to improve human visual perception and sharpen the geometry and color boundaries, we applied the two adversarial losses, $L^{adv}_{color}$ and $L^{adv}_{edge}$, from our dual-stream discriminator along with the conventional content loss $L_{cont}$~\cite{gatys2016image}. The overall loss is formulated as below:
% \vspace{-0.2cm}
% \begin{equation}
% L = \lambda_{1}L_{geo} + \lambda_{2}L_{color} + L_{depth} + \lambda_{3}L_{cont} + \lambda_{4}(L^{adv}_{color} + L^{adv}_{edge}),
% \label{eq:loss}
% \vspace{-0.2cm}
% \end{equation}
% where $\lambda_1$, $\lambda_2$, $\lambda_3$, $\lambda_4$ are the scaling coefficients.

% 10/04 Ruby
\vspace{-0.2cm}
\subsection{Training Loss} %include problem and input definition
\label{sec:M_train_loss}

%Our training objectives can be categorized into two groups. For the naive full supervision loss, we supervise our geometry (the predicted TSDF scene $T_{y}$ to the target data $T_{t}$ in 3D space) and color reconstruction (the rendered color $C_{y}$ and depth $D_{y}$ images to the target images $C_{t}$ and $D_{t}$ in 2D space). The $L_{geo}$ following the log-transformed $\ell1$ TSDF loss in~\cite{dai2020sg} (compute on the predicted TSDF locations c with a total number of locations $N_{c}$) is applied to penalize the geometry in 3D space, and the $L_{color}$ as well as depth $L_{depth}$ (both are $\ell1$ losses and are operated only on valid pixels p with a total number of valid pixels $N_{p}$) are leveraged to ensure geometry and color reconstruction on the rendered images.

Our training objectives can be categorized into two groups, extended from~\cite{dai2020sg} and~\cite{dai2021spsg}. For the naive full supervision loss, we supervise our geometry (the predicted TSDF scene $T_{y}$ to the target data $T_{t}$ in 3D space) and color reconstruction (the rendered color $C_{y}$ and depth $D_{y}$ images to the target images $C_{t}$ and $D_{t}$ in 2D space). The log-transformed $\ell1$ TSDF loss (computed on the predicted TSDF locations $c$ with a total number of locations $N_{c}$), denoted as $L_{geo}$, is applied to penalize the geometry in 3D space. The $L_{color}$ as well as depth $L_{depth}$ (both are $\ell1$ losses and are operated only on valid pixels $p$ with a total number of valid pixels $N_{p}$) are leveraged to ensure geometry and color reconstruction on the rendered images.

% \vspace{-0.2cm}
\begin{equation}
%L = \lambda_{1}L_{geo} + \lambda_{2}L_{color} + L_{depth},
L_{geo} = \frac{1}{N_{c}} \sum_{c} ||sign(T_{y}(c)) \cdot log T_{y}(c) - sign(T_{t}(c)) \cdot log T_{t}(c)||_{1}
\label{eq:loss_geo}
\vspace{-0.2cm}
\end{equation}

\begin{equation}
%L = \lambda_{1}L_{geo} + \lambda_{2}L_{color} + L_{depth},
L_{color} = \frac{1}{N_{p}} \sum_{p} || C_{y}(p) - C_{t}(p)||_{1} \quad L_{depth} = \frac{1}{N_{p}} \sum_{p} || D_{y}(p) - D_{t}(p)||_{1}
\label{eq:loss_color}
\vspace{-0.2cm}
\end{equation}

%Furthermore, to improve human visual perception and sharpen the geometry and color boundaries, we applied the two conditional adversarial losses, $L^{adv}_{color}$ and $L^{adv}_{edge}$ on the rendered color $C_{y}$, normal $N_{y}$, and edge $E_{y}$ images, from our dual-stream discriminator along with the conventional content loss $L_{cont}$~\cite{gatys2016image}. $[\cdot, \cdot]$ means concatenation. $x_{cn} = [C_{x}, N_{x}]$ and $E_{x}$ is the rendered images from input data.

Furthermore, to improve visual quality and sharpen the geometry and color boundaries, apart from the conventional content loss $L_{cont}$~\cite{gatys2016image}, we applied two conditional adversarial losses ($L^{adv}_{color}$ and $L^{adv}_{edge}$) on the rendered color $C_{y}$, normal $N_{y}$, and edge $E_{y}$ images. $[\cdot, \cdot]$ means concatenation. $x_{cn} = [C_{x}, N_{x}]$ and $x_{e} = E_{x}$ is the rendered images from input data.

\vspace{-0.2cm}
\begin{equation}
L^{adv}_{color} = \mathbb{E}_{x_{cn},C_{y},N_{y}}(\log D(x_{cn},[C_{y},N_{y}])) + \mathbb{E}_{x_{cn},C_{t},N_{t}}(\log (1-D(x_{cn}, [C_{t},N_{t}])))
%L^{adv}_{color}=\mathbb{E}_{C_{x},N_{x},C_{y},N_{y}}(\log D([C_{y\x},N_{x}],[C_{y},N_{y}]))+\mathbb{E}_{C_{y\x},N_{x},C_{t},N_{t}}(\log (1-D([C_{y\x},N_{x}], [C_{t},N_{t}])))
\label{eq:loss_adv_color}
\vspace{-0.2cm}
\end{equation}

\vspace{-0.2cm}
\begin{equation}
L^{adv}_{edge} = \mathbb{E}_{x_{e},E_{y}}(\log D(x_{e},E_{y})) + \mathbb{E}_{x_{e},E_{t}}(\log (1-D(x_{e}, E_{t})))
\label{eq:loss_adv_edge}
\vspace{-0.2cm}
\end{equation}

\vspace{-0.2cm}
\begin{equation}
L_{cont} = ||\textrm{VGG}_8(C_{y}) - \textrm{VGG}_8(C_{t})||_{2}
\label{eq:loss_cont}
\vspace{-0.2cm}
\end{equation}

The overall loss is formulated as below, and $\lambda_1$, $\lambda_2$, $\lambda_3$, $\lambda_4$ are the scaling coefficients:

\vspace{-0.2cm}
\begin{equation}
L = \lambda_{1}L_{geo} + \lambda_{2}L_{color} + L_{depth} + \lambda_{3}L_{cont} + \lambda_{4}(L^{adv}_{color} + L^{adv}_{edge}).
\label{eq:loss}
\vspace{-0.2cm}
\end{equation}
\section{Experiments}
\label{sec:exp}

\subsection{Experimental Settings} %evaluation metrics
\label{sec:E_train}

{\bf Training Settings:} Our network is trained on a single NVIDIA GeForce RTX 2080 Ti with a batch size of 2, and it takes about 6 epochs $\approx$ 48 hours to train until convergence. It is optimized via an Adam optimizer with a learning rate of 0.0001. The patch size of the discriminator is 94 x 94 cropped from 320 x 256 images. The $\lambda_1,\lambda_2,\lambda_3,\lambda_4 $ in Eq.~\ref{eq:loss} are set as 0.3, 0.6, 0.01 and 0.005 via grid search. In the training stage, we crop the 3D scene into 64 x 64 x 128 chunk voxels to speed up the process. In the testing stage, we directly input the room-sized scene to our model as 3D CNN is invariant to the scene scale.

\noindent {\bf Evaluation Metrics:} For a fair comparison, we follow the evaluation metric in~\cite{dai2021spsg}. The geometry performance is evaluated by IoU, Recall, and Chamfer Distance. Note that only the observed regions in the target scene are evaluated, and we ignore the unobserved areas as the same in~\cite{dai2021spsg}. Besides, the color performance is evaluated by SSIM (structural similarity image metric)~\cite{brunet2011mathematical}, Feature-$\ell1$~\cite{oechsle2019texture}, and FID (Fr\'echet Inception Distance)~\cite{heusel2017gans} to capture the differences at both local and global scales between the rendered and target images.

\noindent {\bf Baselines:} To verify the proposed dual-stream GAN in our novel task, we compare it with several SOTA 3D completion approaches, including PIFu$^+$~\cite{saito2019pifu}, SG-NN~\cite{dai2020sg}, and SPSG~\cite{dai2021spsg}. Also, we develop another baseline, SPSG equipped with mask inputs, to validate the importance of mask information. We follow \cite{dai2021spsg} to implement all baselines and exclude SG-NN from the evaluation of color performance as it is designed to complete geometry only.

\vspace{-0.2cm}
\setlength{\tabcolsep}{0.011\linewidth}{
\begin{table}[h!]
\centering
% \tiny

% \resizebox{\textwidth}{!}{
\small
\begin{tabular}{c|ccc|ccc}
\toprule
\multirow{2}{*}{\begin{tabular}[c]{@{}c@{}}Methods\end{tabular}} & \multicolumn{3}{c|}{Geometry} & \multicolumn{3}{c}{Color}\\
& IoU($\uparrow$)    & Recall($\uparrow$)    & CD($\downarrow$)  & SSIM($\uparrow$)  & Feature-$\ell1$($\downarrow$)  & FID($\downarrow$)  \\
\midrule
PIFu$^+$ \cite{saito2019pifu}       & 0.241 & 0.525 & 19.537 & 0.744 & 0.253 & 108.87 \\
SG-NN \cite{dai2020sg}              & 0.636 & 0.857 & 20.988 & - & - & - \\
SPSG \cite{dai2021spsg}     & 0.466 & 0.683 & 17.457 & 0.829 & 0.220 & ~~75.10 \\
SPSG (+mask) \cite{dai2021spsg} & 0.491 & 0.659 & ~~3.336 & 0.843 & 0.214 & ~~69.60 \\
\midrule
\bf{Ours}                           & \bf{0.781} & \bf{0.896} & \bf{~~2.110} & \bf{0.853} & \bf{0.209} & \bf{~~65.28} \\
\bottomrule
\end{tabular}
% }
\vspace{0.2cm}
\caption{The comparison of 3D geometric and color inpainting performance on the FF-Matterport. With mask information and proper color and geometry interaction, our dual-stream GAN outperforms all the baselines in both geometry and color metrics. Notably, models without masks all fail in the CD metric due to generating redundant meshes outside the missing areas, which verifies the importance of masks on the 3D scene inpainting task.}
\vspace{-0.6cm}
\label{Tab:baseline}
\end{table}
}
\setlength{\tabcolsep}{0.011\linewidth}{
\begin{table}[h!]
\centering
% \tiny

% \resizebox{\textwidth}{!}{
\small
\begin{tabular}{c|ccc|ccc}
\toprule

\multirow{2}{*}{\begin{tabular}[c]{@{}c@{}}Methods\end{tabular}} & \multicolumn{3}{c|}{Geometry} & \multicolumn{3}{c}{Color}\\
& IoU($\uparrow$)    & Recall($\uparrow$)    & CD($\downarrow$)  & SSIM($\uparrow$)  & Feature-$\ell1$($\downarrow$)  & FID($\downarrow$)  \\
\midrule
\bf{Dual-stream GAN (Full)}    & \bf{0.781} & \bf{0.896} & \bf{~~2.110} & \bf{0.8536} & \bf{0.209} & \bf{65.28} \\
- Edge Discriminator           & 0.774 & 0.892 & ~~2.137 & 0.8534 & \bf{0.209} & 65.62 \\
- 3D GatedConv                 & 0.747 & 0.875 & ~~2.250 & 0.8491 & 0.211 & 68.28\\
- Mask Info.                   & 0.592 & 0.827 & 24.356 & 0.8471 & 0.213 & 69.75\\
\midrule
Single-stream GAN              & 0.744 & 0.878 & ~~2.523 & 0.8511 & 0.210 & 66.68\\
\bottomrule
\end{tabular}
% }
\vspace{0.2cm}
\caption{The ablation studies of 3D geometric and color inpainting performance on the FF-Matterport. Compared with the single-stream GAN, our dual-stream GAN reaches higher performance in all metrics, especially the three geometry metrics. Also, we show the effectiveness of 3D GatedConv and mask components on the 3D scene inpainting task.}
\vspace{-0.6cm}
\label{Tab:ablation}
\end{table}
}

\subsection{Main Results} % geo & color and quality & quantity tables/images
\label{sec:E_result}

\begin{figure*}[t]
\vspace{-0.1cm}
\begin{center}
\includegraphics[scale=0.3]{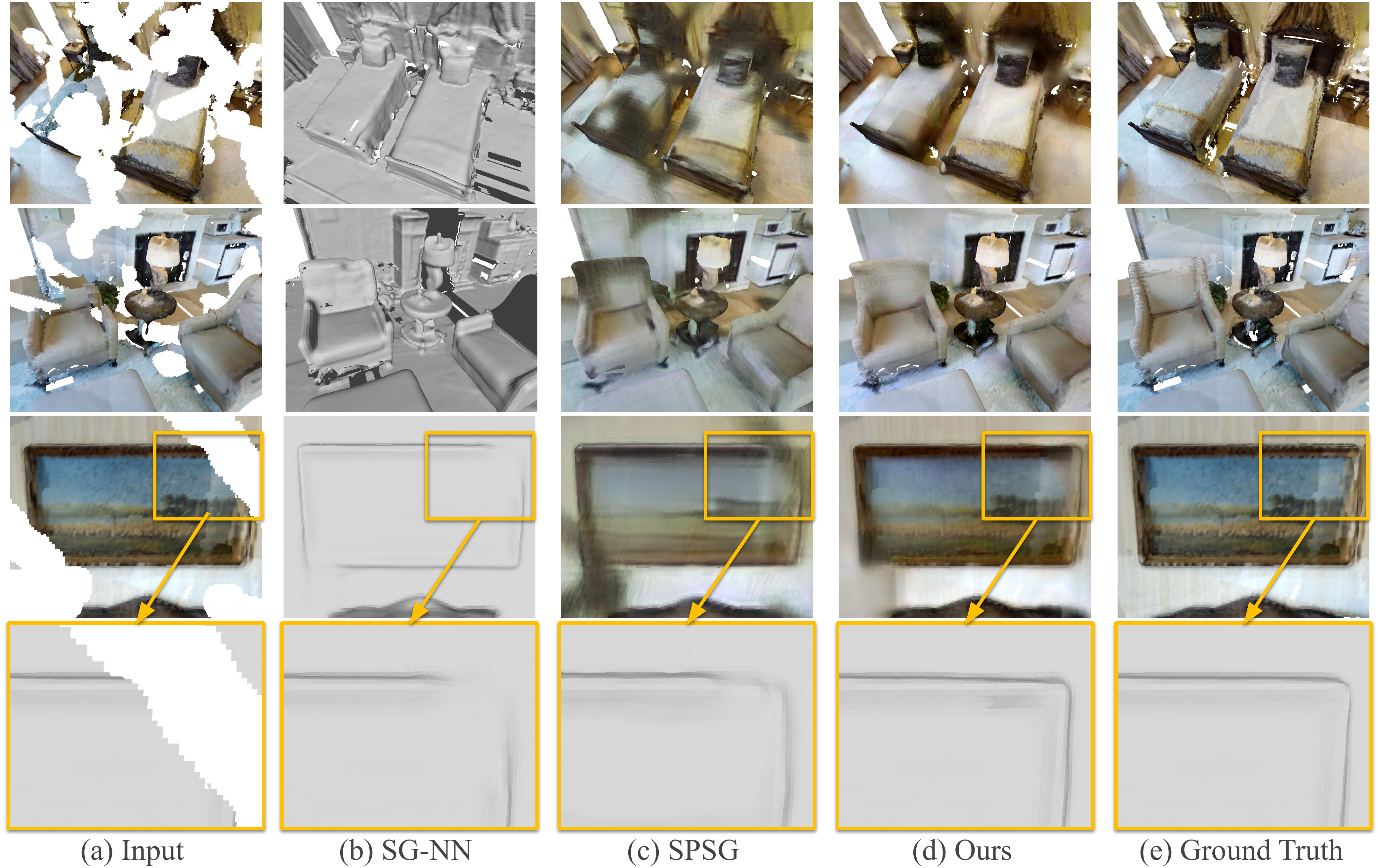}
\vspace{-0.6 cm}
\end{center}
   \caption{Qualitative results of all methods on the FF-Matterport (Best viewed in zoomed digital). Compared to our dual-stream GAN, SPSG fails to preserve details from input and SG-NN predicts distorted structures with small holes in large missing areas in the 1\ts{st} and 2\ts{nd} rows. To further show the fine details of our predicted mesh, the 4\ts{th} row zoom in on the missing corner of the 3\ts{rd} row. More analyses and discussions are reported in Sec.~\ref{sec:E_result}.
   }
\vspace{-0.2 cm}
\label{fig:result}
\end{figure*}

We verify all methods on the ~\ourdataset{} dataset and illustrate the quantitative and qualitative results in Tab.~\ref{Tab:baseline} and Fig.~\ref{fig:result}, respectively. From Tab.~\ref{Tab:baseline}, PIFu$^+$ obtains the worst performance in most metrics, indicating that it cannot adapt to tackle this novel task. About SG-NN, it captures more local features, resulting in higher IoU and Recall scores than other baselines. But it fails to capture the global features and generates distorted structure with small holes in large missing areas, such as the corner of the bed (1\ts{st} row) and the sofa chair on the left (2\ts{nd} row) of Fig.~\ref{fig:result} (b). Thus, it leads to bad performance on the CD metrics. 

Regarding SPSG, the 1\ts{st} row in Fig.~\ref{fig:result} (c) shows that it fails to preserve the details of complete parts from input scenes, such as the lamp beside the sofa, and results in lower IoU and Recall scores. Besides, SPSG suffers from initial color bias pointed in \cite{yu2019free} and performs poorly on the color scores. On the contrary, our dual-stream GAN better utilizes the mask information as well as both structure and semantic feature from dual-stream to generate well-structured and semantic reasonable 3D scenes; accordingly, we achieve the highest performance in all metrics.

To show the further restoration details, the 4\ts{th} row of Fig.~\ref{fig:result} zooms in on the missing corner of the frame in the 3\ts{rd} row. Unlike SG-NN and SPSG flattening and ignoring the frame structure, our model easily distinguishes the frame from the wall using color features and recovers the corner with straight and delicate edges guided by our edge adversarial loss.

\subsection{Ablation Studies}
\label{sec:E_abla}

We summarize the ablation study in Tab.~\ref{Tab:ablation}. First, we analyze the influence of our edge discriminator. Comparing the 1\ts{st} and 2\ts{nd} rows, the edge discriminator causes minor improvements in numerical evaluations but significantly contributes to the visualization (in supplementary). This phenomenon echoes our hypothesis that 2D edge loss can guide 3D geometry and color to better collaborate on generating delicate 2D edges. Then, we verify the design of our pipeline. We build a single-stream sequential generator with mask inputs and 3D GatedConv, named single-stream GAN in Tab.~\ref{Tab:ablation}. The single-stream GAN generates geometry features without knowing color features and then passes them to the color stream. As a result, its geometry performance declined larger than color performance.

Lastly, we examine the efficacy of 3D GatedConv and mask information. As shown in the 3\ts{rd} and 4\ts{th}rows of Tab.~\ref{Tab:ablation}, they substantially improve the performance, both in geometry and color scores, which reveals the mask information is indispensable in the novel 3D inpainting task. This observation is consistent with the advancement of SPSG with mask inputs in the main experiment (Tab.~\ref{Tab:baseline}). Especially the CD metric is evaluated on the fixed-number points uniformly sampled from predicted meshes, models without mask information tend to generate redundant meshes outside the missing areas, resulting in bad performance. Through these analyses, we demonstrate the importance of mask information on the challenging 3D scene inpainting task, which means the proposed dataset and approach are requisite.

%10/03 Ruby
\vspace{-0.2cm}
\subsection{Practical Application: Object Removal}
\label{sec:E_objectRemove}

\begin{figure*}[h]
%\vspace{-0.35cm}
\begin{center}
\includegraphics[scale=0.3]{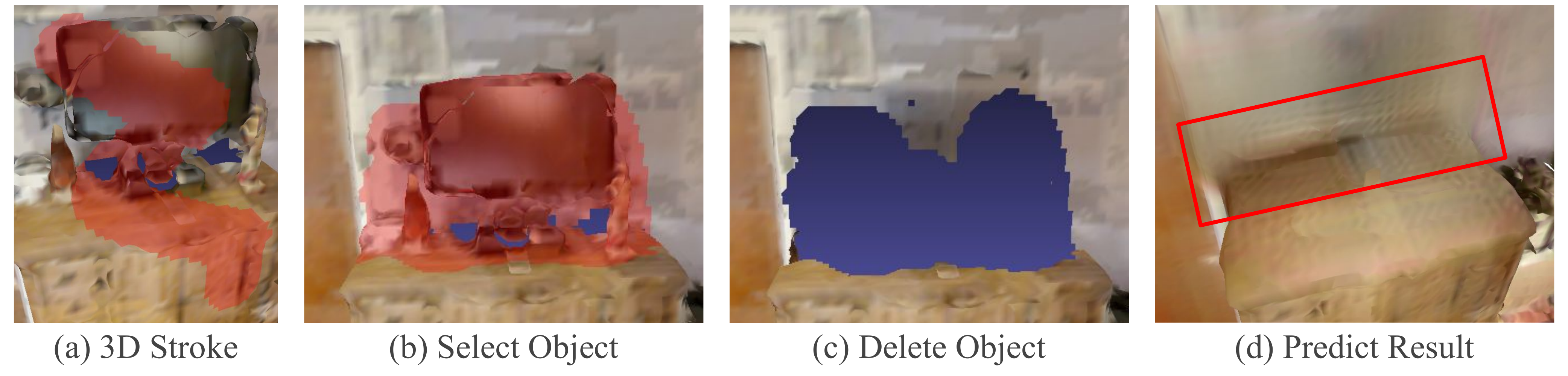}
\vspace{-0.75cm}
\end{center}
   \caption{An example of object removal. (a), (b), and (c) show the Z-painting tool in MeshLab can draw 3D strokes to select and delete the unwanted area in a scene. Our predicted result is shown in (d).}
\vspace{-0.2cm}
\label{fig:real}
\end{figure*}

Our 3D scene inpainting task is designed for user editing purposes, like removing unwanted objects when constructing realistic scenes, practical for AR/VR applications. To further demonstrate the whole process of scenes application, we apply our method to an example of 3D object removal with real manual masks in Fig.~\ref{fig:real}. With existing 3D editing tools (e.g. MeshLab\footnote {https://www.meshlab.net/}), (a) users can easily draw strokes with paintbrushes in 3D scenes to (b) select unwanted regions like the screen and (c) mask out them. Given the remained scene with the masked regions (red areas in (b)), our model can fill these regions with realistic contexts (e.g. the sharp corner marked in red) (As shown in (d)).

\section{Conclusion}
\label{sec:con}

In this paper, we propose a novel free-form 3D scene inpainting task with a new FF-Matterport dataset imitating user drawing masks in real-world 3D editing applications. We further introduce a dual-stream GAN model to address the increased difficulty of the new task and achieve SOTA results. To elaborate, our dual-stream generator utilizes both geometry and color information to recover the semantic boundaries in large missing regions, and our dual-stream discriminator enhances the inpainted details to be realistic and delicate. % Experiments on FF-Matterport dataset show that our method outperforms the existing scene completion methods on both qualitative and quantitative results.

\section*{Acknowledgement}

%This work was supported in part by the National Science and Technology Council under Grant MOST 110-2634-F-002-051, Mobile Drive Technology Co., Ltd (MobileDrive), and NOVATEK fellowship. We are also grateful to the National Center for High-performance Computing.

This work was supported in part by the National Science and Technology Council, under Grant MOST 110-2634-F-002-051, Mobile Drive Technology Co., Ltd (MobileDrive), and NOVATEK fellowship. We are also grateful to the National Center for High-performance Computing.

% \bibliography{egbib}
\bibliography{arxiv}
\clearpage

% supplementary
\appendix

\title{Supplementary Material}
\maketitlesup

% \vspace{2cm}
% \section*{Supplementary Material}

\begin{abstract}

% 10/04 Ruby
In this supplementary material, we detail our free-form 3D mask generation algorithm for the FF-Matterport dataset in Sec.~\ref{sec:freeform} with pseudo code in Algorithm~\ref{alg:free-form}. We then provide the implementation details of our designed network dual-stream generator and the NN-based edge detector in dual-stream discriminator in Sec.~\ref{sec:arch}. Furthermore, we show additional results in  Sec.~\ref{sec:addresult}, including the qualitative (perceptual) improvement of our edge discriminator (Sec.~\ref{sec:edgedisc}) and the limitations of our proposed model (Sec.~\ref{sec:limit}).

\end{abstract}

% \input{final_supp_sec/1_arxiv_FreeForm}
% \clearpage
% \input{final_supp_sec/2_NetworkArchitecture}
% \input{final_supp_sec/3_arxiv_AddResult}
\section{Free-form 3D Mask Generation Algorithm}
\label{sec:freeform}

\algnewcommand\algorithmicforeach{\textbf{for each}}
\algdef{S}[FOR]{ForEach}[1]{\algorithmicforeach\ #1\ \algorithmicdo}

\begin{algorithm}[!h]
    \begin{algorithmic}[1]
    \setstretch{1.3}
    \State \textbf{Input:} Original scene: $S_{o} = \{T_{o}, C_{o}\}$, and $diameter$, $maxStrokeStep$, $totalStep$.
    \State \textbf{Output:} Masked scene: $S_{m} = \{T_{m}, C_{m}, M_{m}\}$.
    \State \textbf{Init:} Copy from original scene $T_{m} \leftarrow T_{o}$, $C_{m} \leftarrow C_{o}$, $(\#\textrm{step}, \#\textrm{strokeStep}) \leftarrow (0, 0)$.
    % \Function{RandomCenter}{$T, t$}\Comment{T is TSDF values and t is occupancy threshold}
    %     \State \Return $\textrm{random.choice}(|T| \leq t)$
    % \EndFunction
    \Function{RandomMaxStroke}{$s$}
        \State \Return random.randint(s, s+10)
    \EndFunction
    \Function{FindValidBall}{$c, d, t$}\Comment{c is center, d is diameter, and t is threshold}
        \State \Return $X \leftarrow \{x|x \in B_{c,d} \cap |T_{m}(x)| \leq t \}$\Comment{$B_{c,d}$ is a ball centered at c with diameter d}
    \EndFunction
    \State $\textrm{center}\ O_c \leftarrow \textrm{random.choice}(|T_{m}| \leq 1)$\Comment{Random a starting point on occupied voxels}
    % \State $\textrm{center}\ O_c \leftarrow$ \Call{RandomCenter}{$T_{m}$}\Comment{Random a starting point on occupied voxels}
    % \State $\textrm{center}\ O_c \leftarrow$ $\textrm{random.choice}(|T_{i}| \leq 1)$\Comment{Random a starting point on occupied voxels}
    \State $\textrm{max stroke step}\ L \leftarrow \Call{RandomMaxStroke}{maxStrokeStep}$
    % \State $\textrm{max stroke step}\ L \leftarrow random.randint(maxStrokeStep, maxStrokeStep+10)$
    \While{$\#\textrm{step} \leq TotalStep$}
        \State Mask out $B_{O_c, diameter}$ in $\{T_{m}, C_{m}, M_{m}\}$\
        \State $\#\textrm{step} \leftarrow \#\textrm{step} + 1$
        \State $\#\textrm{strokeStep} \leftarrow \#\textrm{strokeStep} + 1$
        \If {$\#\textrm{strokeStep} \geq L$}\Comment{Restart a new stroke}
            % \State $\textrm{center}\ O_c \leftarrow$ \Call{RandomCenter}{$T_{i}$}
            \State $\textrm{center}\ O_c \leftarrow$ $\textrm{random.choice}(|T_{m}| \leq 1)$
            \State $\textrm{max stroke step}\ L \leftarrow \Call{RandomMaxStroke}{maxStrokeStep}$
            \State $\#\textrm{strokeStep} \leftarrow 0$
        \ElsIf {$X \leftarrow \Call{FindValidBall}{O_c, diameter//2, 1} \neq \emptyset$}\Comment{Move a small step}
            \State $O_c \leftarrow \textrm{random.choice}(X)$
        \ElsIf {$X \leftarrow \Call{FindValidBall}{O_c, diameter, 5} \neq \emptyset$ }\Comment{Move a big step}
            \State $O_c \leftarrow \textrm{random.choice}(X)$
        \Else\Comment{Dead end, Restart a new stroke}
            % \State $\textrm{center}\ O_c \leftarrow$ \Call{RandomCenter}{$T_{i}$}
            \State $\textrm{center}\ O_c \leftarrow$ $\textrm{random.choice}(|T_{m}| \leq 1)$
            \State $\textrm{max stroke step}\ L \leftarrow \Call{RandomMaxStroke}{maxStrokeStep}$
            \State $\#\textrm{strokeStep} \leftarrow 0$
        \EndIf
    \EndWhile
    \State \Return $\{T_{m}, C_{m}, M_{m}\}$.
    \caption{Free-form 3D Mask Generation Algorithm}
    \label{alg:free-form}
    \end{algorithmic}
\end{algorithm}

As stated in Sec.\ref{sec:M_data} of the main paper, our designed free-form 3D mask generation algorithm aims to mimic human drawing trajectories in 3D space and randomly generate diverse free-form masks for efficient training and evaluation. To avoid covering the empty space in the scene into the masked areas and to flexibly draw arbitrary shapes of 3D objects, we utilize the characteristic of TSDF and dynamically decide the length and direction of strokes based on the TSDF values. We show our algorithm in Algorithm \ref{alg:free-form}.

The input of the algorithm is an original scene $S_{o}$ from a real-world scanned scene containing $T_{o}$ TSDF and $C_{o}$ color voxelized values, and three hyper-parameters $diameter$, $maxStrokeStep$, and $totalStep$ to control the mask distribution. The output is a masked scene $S_{m}$ for training and inferring consisting of masked $T_{m}$ TSDF and $C_{m}$ color voxelized values with corresponding binary mask map $M_{m}$. This algorithm is for the 64x64x128 chunk size in the training dataset. For the test dataset which contains whole indoor rooms, We run this algorithm several times in equal proportions to the size of the indoor rooms.

\clearpage
\section{Network Architecture}
\label{sec:arch}

\begin{figure*}[!h]
\begin{center}
%\includesvg[scale=0.65]{images/supp_fig_network_architecture_5.svg}
% \includegraphics[scale=0.59]{images/supp_fig_network_architecture_6.pdf}
\includegraphics[scale=0.59]{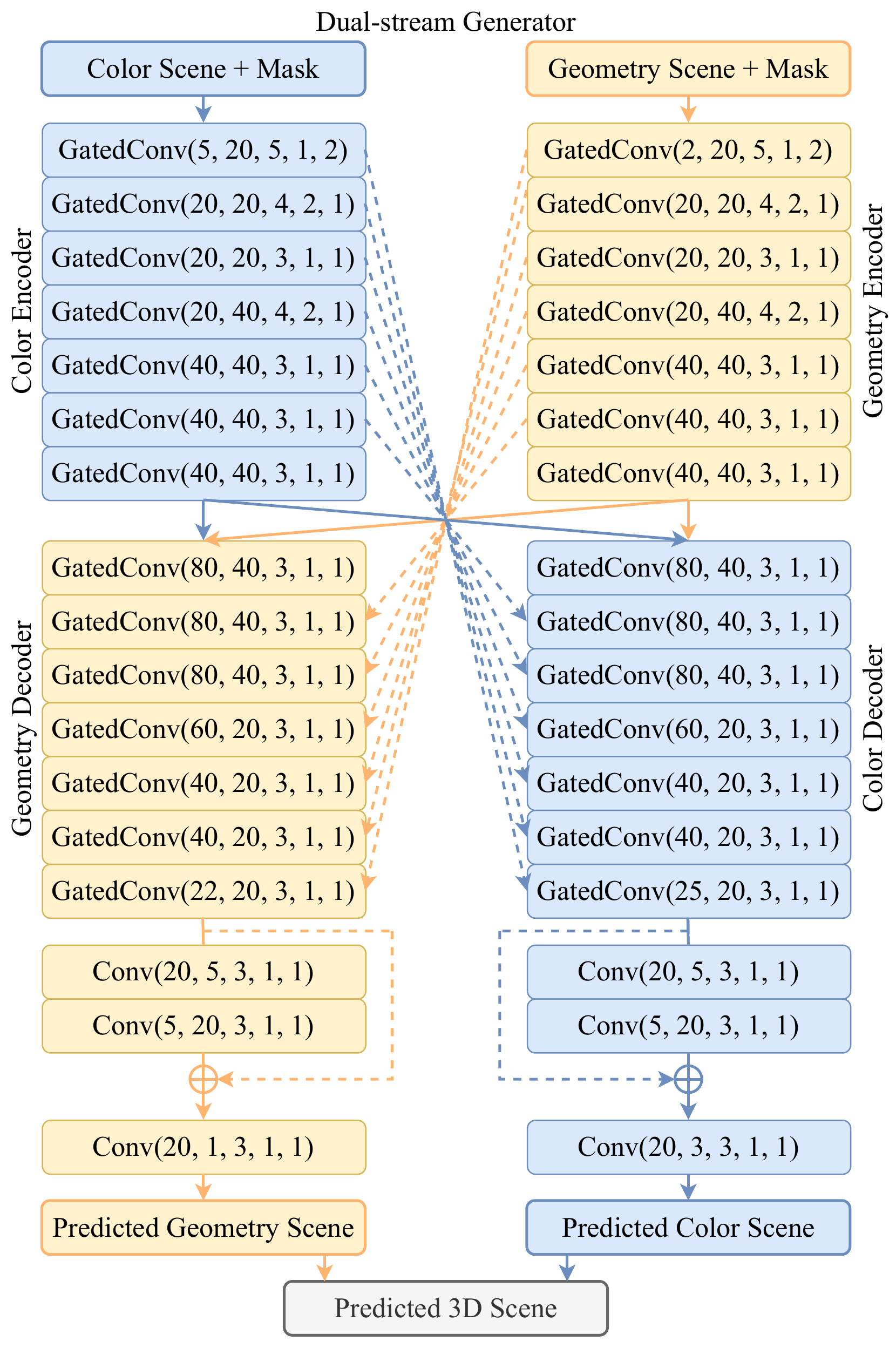}
\vspace{-0.6cm}
\end{center}
   \caption{Details of dual-stream generator architecture, where $\oplus$ denotes add and dashed line denotes skip connection. We process the geometry and color scenes in two streams and concatenate the predicted geometry and color scenes to get the final result.}
\label{fig:fig_arch_dualgen}
\end{figure*}

\begin{figure*}[!h]
\begin{center}
\includegraphics[scale=0.55]{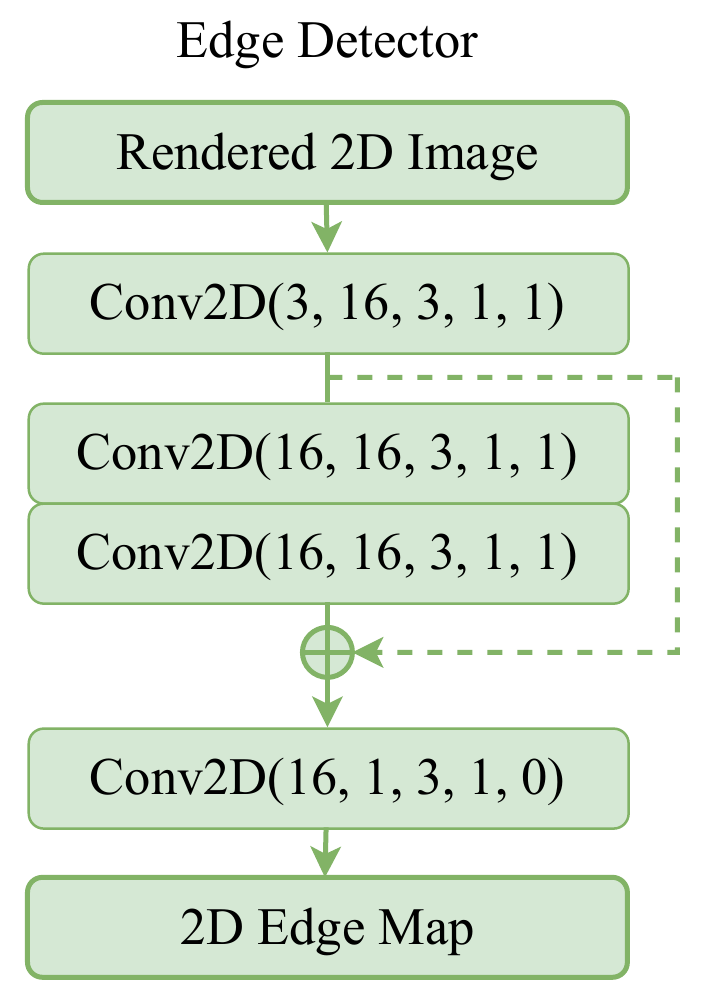}
\vspace{-0.5cm}
\end{center}
  \caption{Details of edge detector architecture of dual-stream discriminator, where $\oplus$ denotes add and dashed line denotes skip connection. The edge detector takes the rendered 2D image from the differentiable 2D rendering and extracts the corresponding edge map for our edge-stream discriminator.}
 \vspace{-0.5cm}
\label{fig:fig_arch_edgedet}
\end{figure*}

We show the implementation details of our proposed network architectures: dual-stream generator in Fig.~\ref{fig:fig_arch_dualgen} and the NN-based edge detector of dual-stream discriminator in Fig.~\ref{fig:fig_arch_edgedet}. The implementation of the patch-based discriminator of dual-stream discriminator follows the one in SPSG~\cite{dai2021spsg}. In the figures, GatedConv and Conv stand for the 3D gated convolutional layer and 3D convolutional layer. Conv2D denotes the 2D convolutional layer. The parameters of convolution are given by (in-channel, out-channel, kernel size, stride, padding), and all GatedConv layers are paired with a 3D batch normalization and a Leaky ReLU.

\vspace{-0.4cm}

\section{Additional Results}
\label{sec:addresult}
%Ablation Qualitative Result of Edge Discriminator

\subsection{Edge Discriminator Qualitative Result}
\label{sec:edgedisc}

\begin{figure*}[!h]
\begin{center}
\includegraphics[scale=0.305]{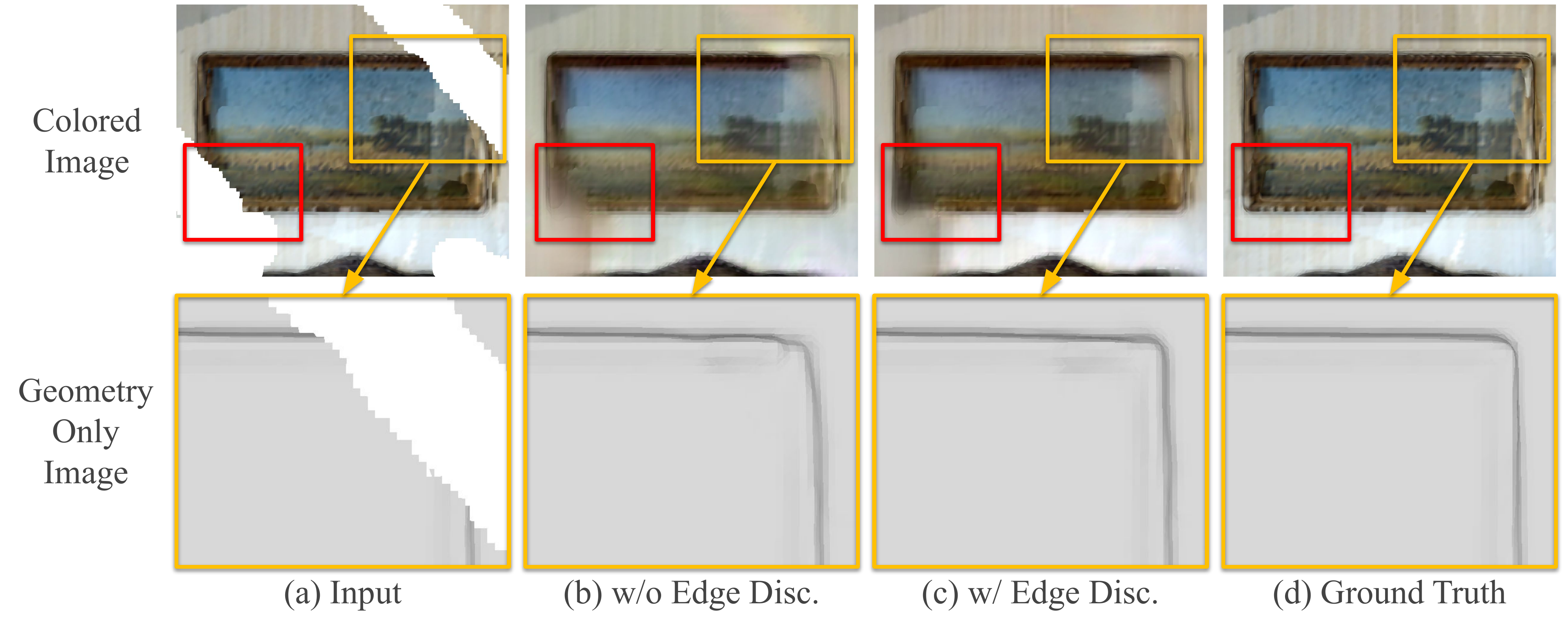}
\vspace{-1cm}
\end{center}
   \caption{Additional qualitative results of the dual-stream discriminator on FF-Matterport. By adding the edge stream to the conventional color stream discriminator, the color boundary becomes less blurred (shown in red frames), and the geometry shapes become more straight and sharper (shown in zoomed-in pictures in the 2\ts{nd} row with yellow frames).}
\label{fig:fig_edgedisc}
\end{figure*}

In the Sec.\ref{sec:M_dual_g} of the main paper, we state that our additional discriminator aims to enhance the sharpness and details of the results for better human perception. As shown in Fig.~\ref{fig:fig_edgedisc}, we provide additional qualitative results on FF-Matterport to demonstrate the efficacy of our dual-stream discriminator. Compared with (b) the model without our edge discriminator, the full model (c) performs better on both geometry and color. In (c), the bottom left corner (red frame) of the color image becomes less blurred after adding the edge stream, and the top right corner (yellow frame) of the color image zoomed in as the geometry only image shows that the geometry shapes become sharper after adding our edge discriminator. This phenomenon echoes our motivation that 2D edge loss can simultaneously guide the 3D geometry and color streams to generate delicate edges.

\subsection{Limitation}
\label{sec:limit}

\begin{figure*}[t]
\begin{center}
\includegraphics[scale=0.305]{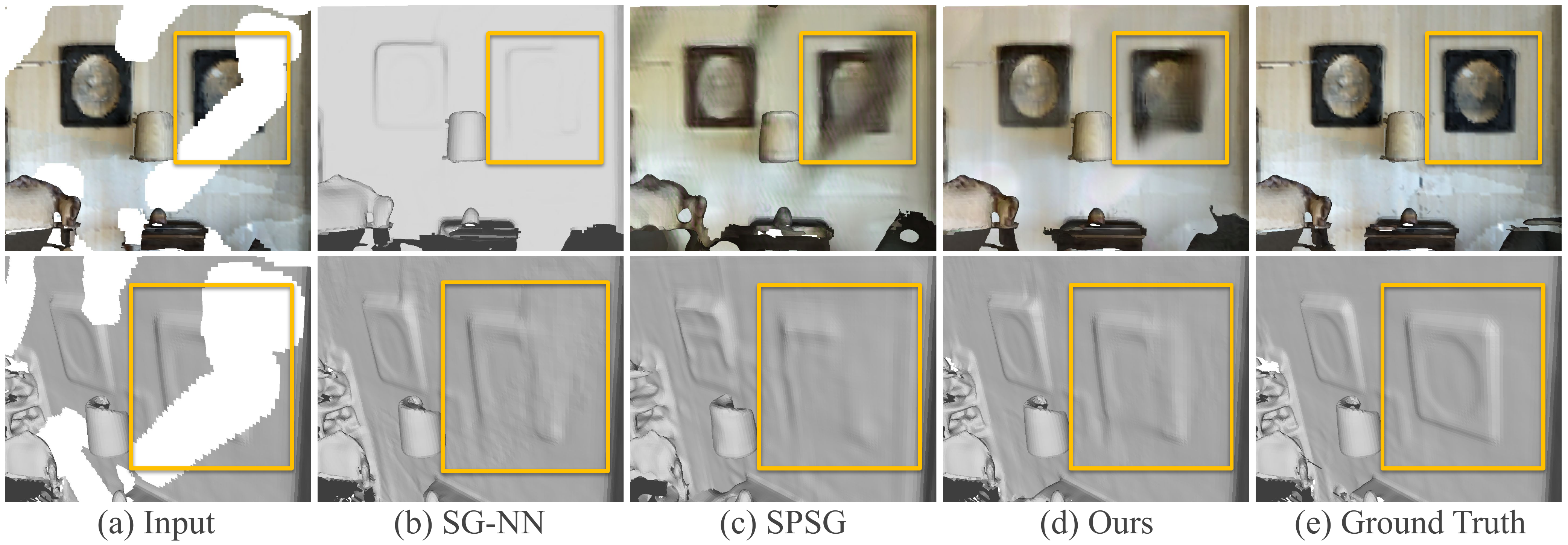}
\vspace{-1.1cm}
\end{center}
   \caption{A failure case of the frame on the right side in (a) losing most of the edges and color. Even though our model fails to predict distinct structures and edges in the missing parts, our results (d) still outperform the baselines SG-NN~\cite{dai2020sg} (b) and SPSG~\cite{dai2021spsg} (c) in both geometry and color visual performance.}
\label{fig:fig_limit}
\end{figure*}

Although our proposed method can generate realistic geometry and color results for the real-world 3D scene inpainting task, we still find its limitations and unsolved challenges as shown in Fig.~\ref{fig:fig_limit}. For example, in the 1\ts{st} row (a), the right side of an incomplete picture frame loses most of the edges and color; also, the 2\ts{nd} row (a) shows that the structure of the frame is very close to the wall and hard to be distinguished. Therefore, even with the help of our dual-stream GAN design, the model fails to predict the correct edges of the top right and bottom left corners, resulting in blurred color boundaries in the comparison of (d) and (e). Still, our predicted results contain more details than  SG-NN~\cite{dai2020sg} (b) and SPSG~\cite{dai2021spsg} (c).

Moreover, due to the natural limit of CNN models and voxel representation, the output resolution is restricted. Even though some new 3D data representations are proposed recently and claimed to support the unlimited resolution, such as implicit function, it is still challenging to properly handle the mask information in the 3D scene inpainting task. 

\end{document}